# Road Damage Detection Acquisition System based on Deep Neural Networks for Physical Asset Management

A.A. Angulo[1], J.A. Vega-Fernández[1], L.M. Aguilar-Lobo[1], S. Natraj[2], G. Ochoa-Ruiz[3]

1 Universidad Autónoma de Guadalajara (Mexico),
Email: {andres.angulo, antonio.vega, lina.aguilar}@edu.uag.mx
2 Vidrona LTD (Edinburgh, United Kingdom)
Email: shailendra@vidrona.com
3 ITESM Campus Guadalajara (Mexico)
Email: gilberto.ochoa@tec.mx, ORCID:0000-0002-9896-8727

**Abstract.** Research on damage detection of road surfaces using image processing techniques has been actively conducted, achieving considerably high detection accuracies. So far, most studies focused only on the detection of the presence or absence of damages; however, in real-world scenarios, road managers from need to clearly understand the type of damage and its extent in order to take effective action in advance or to allocate the necessary resources. Moreover, currently there are few uniform and openly available road damage datasets available, leading to a lack of a common benchmark for road damage detection. Such dataset could be used for a great variety of applications; herein, it is intended to serve as the acquisition component of a physical asset management tool which can aid governments agencies for planning purposes, or by infrastructure maintenance companies, so they can implement predictive maintenance procedures.

In this paper, we make two contributions to address these issues. First, we present a large-scale road damage dataset, which includes a more balanced and representative set of damages not present in previous studies. This dataset is composed of 18,0345 road damage images captured with a smartphone installed on a car, with 45,435 instances road surface damages (linear, lateral and alligator cracks, potholes, and various types of painting blurs). In order to generate this dataset, we obtained images from several public datasets and augmented it with crowdsourced images, which where manually annotated for further processing. The images were captured under a variety of weather and illumination conditions. In each image, we annotated the bounding box representing the location and type of damage and its extent. Second, we trained different types generic object detection methods, both traditional (an LBP-cascaded classifier) and deep learning-based, specifically, MobileNet and RetinaNet, which are amenable for embedded and mobile and implementations with an acceptable performance for many applications. We compared the accuracy and inference time of all these models with others in the state of the art, achieving higher accuracies in all the eight classes present in the dataset introduced by researchers at the University of Tokyo, and in other related works, with a lower inference time.

# 1    Introduction

Research on damage detection of road surfaces using image processing and machine learning techniques has been an active area of research in both developed and in-development countries [1-4]. This is an important issue, as roads are one of the most important civil infrastructures in every country and contribute directly and indirectly to the countries' economies and more importantly, to the well-being and safeness of their citizens. Thus, road maintenance is of paramount importance and many countries have implemented inspection mechanisms and standards to carry out this process.

Nonetheless, both the inspection and journaling processes of road damages remain daunting problems, as government agencies still struggle to maintain accurate and up-to-date databases of such structural damages, making it hard to allocate resources for repair works in an informed manner. The problem is exacerbated as the number of experts that can assess such structural damages is limited, and furthermore, methods typically used to collect data from the field are time-consuming, cost-intensive, require a non-trivial level of expertise, and are highly-subjective and prone to errors. Therefore, both academic endeavors and commercial initiatives have been conducted to facilitate this process, making use of a combination of sophisticated technologies [5]. Most of these approaches combine various sensors (i.e. inertial profilers, scanners), but also imaging techniques, which have demonstrated to be particularly fit for the task. The information gathered by these sensors and cameras can be fed to machine learning algorithms and combined with mobile acquisition systems and cloud computing approaches to automate the inspection process or to create end-to-end solutions.

Such endeavors have demonstrated promising results [6], but many of them have been limited to specific types of road damage classes (cracks, patches, potholes) and relatively small and not sufficiently comprehensive datasets. Furthermore, traditional "AI" systems for road damage classification do not to scale well and are not versatile enough to work in the dire situations usually found by road inspection experts. This has dramatically changed with the advent of the computer vision approaches based on deep learning architectures, which have demonstrated tremendous strides in several computer vision tasks and have slowly made their way into road inspection systems.

Such computer vision-based road detection and classification systems are significantly cheaper than competing technologies and if trained appropriately, they can lead to the implementation of sophisticated and cost-effective acquisition systems. Such solutions could be further enhanced if they are implemented on mobile devices for road damage acquisition and stored on cloud computing technologies for further processing and big data analyses. Such an approach could lead to the creation of geo-localized databases containing the current state of the roads and other civil engineering infrastructures, which can be updated and used for forecasting and planning purposes. This is perfectly aligned with the industry 4.0 and digital transformation paradigms and can be easily implemented as business model that can be easily replicated. For instance, the UK has launched the Digital Roads Challenge to attain this goal.

In this paper, we present an initial approach for a digital asset management tool, geared towards road and street management. Compared to previous works, which concentrated their efforts to specific types damages (potholes [1], cracks [2] and patches [3], to cite a few), our proposal builds upon previous work in this domain, by proposing a large dataset of various types of asphalt damages (longitudinal cracks, alligator cracks, potholes, bumps, patches, among others) which is, to the best our knowledge, one of the most comprehensive in the literature. Some of the structural damages typically encountered in the field are depicted on **Figure 1**; these images represent structural damage and not damages on the infrastructure signs and markings.

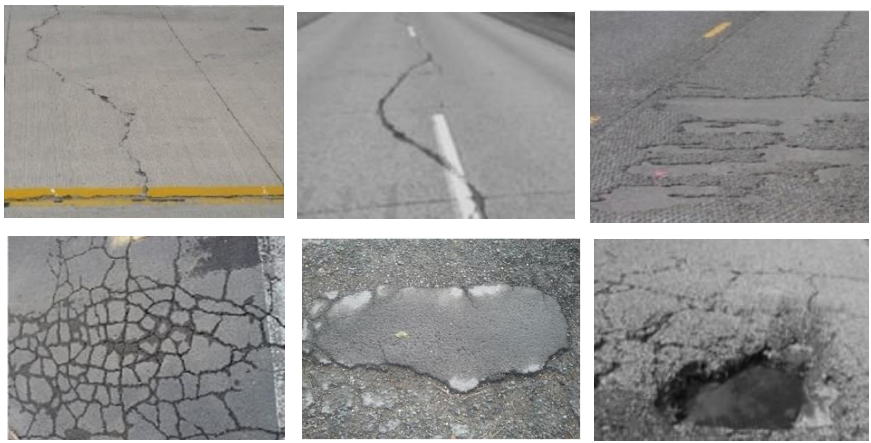

**Fig. 1.** Examples of structural damages typically found in the literature: a-b) linear cracks, c) peeling, d) alligator cracks, e) patches and f) potholes

The proposed road damage dataset has been used to train various machine learning based "object detectors", in which the objects of interest are the structural damages in the dataset. As will be described in more depth in the following section, such systems need to run in real-time to be useful, and therefore, we validated the dataset on light-weight detectors such as LPB cascaded classifiers, and in more recent, deep learning-based Single Shot Detectors (SSD) such as MobileNet and RetinaNet. The two latter were developed so they can run efficiently on mobile devices (i.e. smartphone) in real-time, while achieving high mean average precisions (mAP).

The rest of this paper is organized as follows: in **Section 2** we further motivate the need for digital asset management systems, specifically for roads and streets. Then, in **Section 3** we discuss the state of the art in road damage detection using various imaging modalities to provide some context, but we focus our attention to those techniques based on computer vision and modern machine learning approaches. In **Section 4** we introduce the proposed dataset and an initial approach for road digital asset management, as well as the tools and methods used for validating the proposal. Afterwards, in **Section 5**, we discuss these results with regards with the state of the art and we describe future avenues of research, concluding the article.

## 2 Motivation

The maintenance of roads and street infrastructure is a major concern, especially in tropical countries where the asphalt and other civil facilities are more prone to suffer structural damages. Nevertheless, this problem is not exclusive of these regions, and the issue is becoming pervasive in developed countries due to the aging of their infrastructure. Thus, in order to cope with these problems, governmental agencies in these countries have developed specific procedures and standards to aid experts in assessing the damage on asphalt and other materials such as concrete.

In the field, the experts have traditionally carried inspections based on these standards following two main approaches: either by direct observations by humans or by quantitative analysis using remote sensing approaches. The visual inspection approach requires experienced road managers, which makes it time-consuming, expensive and prone to errors. Moreover, traditional methods tend to be inconsistent and limited to point observations, making it difficult to scale to large cities or geographical areas, and furthermore, to build and maintain digital corpuses with the acquired data.

The problem is exacerbated as the acquisitions made by traditional inspections are not digitally stored, and thus government bodies cannot leverage the power of the digital revolution (i.e. big data analytics) to address these challenges in a more informed and efficient manner. In contrast, more recent quantitative systems, based on large-scale inspection systems using the so-called Mobile Measurement Systems (MMS) or laser-scanning methods have been proposed and deployed in some countries [5].

An MMS can obtain highly accurate geospatial road information using a mobile vehicle; this system comprises a global positioning system (GPS) unit, an internal measurement unit (IMU), digital measurable images obtained through digital cameras, a laser scanner and LIDAR. Although this quantitative inspection systems are highly accurate, they are considerably more expensive and complex to operate, as well as more difficult to acquire, manage, and maintain by small municipalities that lack the required financial resources.

In order to alleviate the above-mentioned problems, a great deal of research efforts have been carried out to facilitate and automate the manual inspection process described above. For instance, the civil engineering domain has a long history of using non-destructive sensing and surveying techniques, typically based on computer vision, such as 3D imaging [6] and remote sensing approaches [7] for a variety of applications. The former approach has several advantages over the traditional inspection methods, as well as to other approaches which use solely conventional imagining techniques, as they cope better with changes in illumination and weather conditions.

Although such tools provide an opportunity for frequent, comprehensive, and quantitative surveys of transportation infrastructures, they are costly and difficult to implement and maintain, as such systems typically rely on heavily instrumented vehicles, as depicted on **Figure 2**. Therefore, such solutions are is simply not affordable by many countries and even less so for individual municipalities. Thus, researches have sought other methods to reduce the complexity and costs associated with surveying and inspecting road infrastructures, well documented in the literature [7,8].

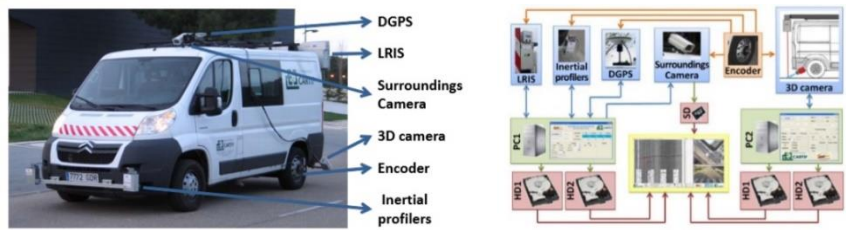

**Fig. 2.** Example of MMS system, designed by PaveMetrics, analyzed in [5]

As we will see in the next section, such efforts have mainly revolved around image processing and computer vision techniques, as the main information used by inspectors can be obtained from conventional imaging when dealing only with classification problems (and not properly surveying), which have made major strides in this domain due to the combination of efficient and lightweight deep learning-based generic object detectors and mobile computing technologies for inference purposes.

## 3       Related Works

The automation of road damage inspection generally requires robust computer vision algorithms with a high degree of intelligence, which can be easy to use and run in real-time. However, most the systems in this vein were initially conceived as means for aiding the experts in the field and were limited by the capabilities of the ML algorithms of the time, which were not very reliable and robust, and which were difficult to implement in portable systems. For a thorough state of the art of the various imaging modalities the reader is directed to excellent surveys in 3D imaging [7], remote sensing [8] and computer vision and image processing in general [9, 10].

The biggest challenge for automated road damage detection systems is to consistently achieve high performance, in terms of accuracy, under various complex environments (due to changes in illumination, weather conditions, among other challenges). Despite these problems, several systems for detecting individual structural damages have been proposed in the literature, as described in **Section 1**. However, most of these works have not been able to tackle more challenging scenarios in the development of their machine learning and AI systems, due to the absence of commonly          accepted and publicly available datasets. The Kitti dataset  [4] could be exploited for such purposes, as it contains thousands of examples of road damages; however, its main purpose is for autonomous driving and using it for road damage detection might prove itself a difficult and ultimately futile endeavor.

Therefore, it was widely recognized by the academic community that there was a need for an extensive dataset specifically designed for road damage assessment, for two main purposes. First, in order to test and perform benchmarks among competing solutions, helping to boost the R&D in the area. Second, to homogenize the data collection process, as such datasets are essential for creating reliable and robust machine learning-based solutions for road and asphalt detection and inspection. This problem has been partially addressed, with some datasets having made available, but much work needs to be done, as roads vary dramatically among countries.

Despite these issues, the available datasets have been exploited over the years to implement machine-learning based solutions, using computer vision-based features and classification algorithms such as SVM for the detection of cracks [11] and potholes [12]. As datasets have become larger, there has been a growing interest in deploying deep learning-based (DL) approaches to the problem of road and asphalt damage detection. It well known [13] that DL algorithms have demonstrated impressive results in a variety of computer vision-related fields and are behind the autonomous driving revolution. In this sense, several efforts have been conducted to leverage these capabilities in the problem we are tackling, with some very impressive results for certain classes of asphalt damages such as cracks [14, 15].

One advantage of deep learning methods over traditional approaches is that cheaper or less sophisticated imaging devices (i.e. smartphones) can be used for acquiring the training samples, instead of the much more expensive MMS platforms described in the previous section. Approaches based on mobile devices had been explored before, using more traditional computer vision detectors such as LBP-cascade classifiers [16, 17], but the authors did not compare their results to more recent detectors based on DL. Furthermore, although the results were commendable, they made use of custom datasets and were limited to more "static analyses" (they could not be run at real time in very complex scenarios), **Figure 3** shows a couple of examples of structural damage detection using the LPB-based approach on OpenCV, using our dataset.

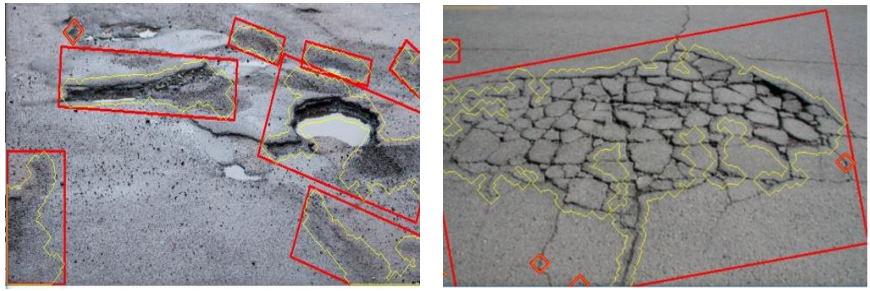

**Fig. 3.** Detection of alligator cracks and potholes using the most basic ML algorithm explore in this work: an LPB-cascade classifier, note the high number of FP.

On the other hand, recent advances in generic object detection and classification algorithms [18] have now made possible to implement very sophisticated and resource efficient DL algorithms in constrained devices, such as mobile phones and smart cameras, and some of the recent implementations have started to emerge in various areas and in particular in the field of road inspection and assessment. In this sense, it is also possible now to carry out the deployment phase (acquisition and detection of structural damages) in real time using inexpensive mobile devices as well [20-21], either for individual inspections in situ or mounted on a car, making it an attractive alternative to the expensive MMS platforms.

These advances have been possible due to the development of two-stage detectors such as Fast R-CNN and subsequently single stage detectors like YOLO (for a detailed discussion the reader is directed to an excellent survey [18]). The former category can achieve high accuracies (mean average precision or mAP) but are usually inefficient, as they cannot be deployed in resource constrained devices [19]. The latter category of generic object detectors has steadily improved over time, with new architectures based on Feature Pyramid Networks (i.e. RetinaNet and MobileNet) achieving satisfactory accuracies for challenging problems whilst requiring a relatively small footprints, leading to implementations in mobile phones for detecting road damages and collecting georeferenced information [19, 21], as depicted on **Figure 4** a) and b)

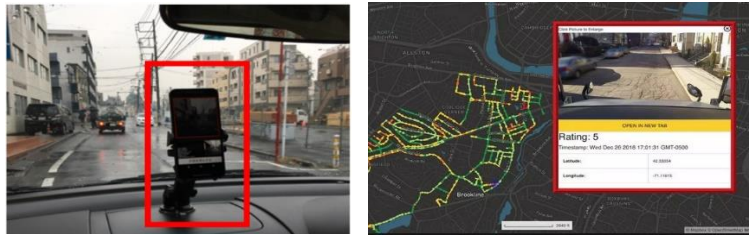

**Fig. 4.** Examples of road damage detection systems running on a smartphone.
a) System by the University of Tokio [21] and b) Solution by RoadBotics [22]

These progresses have also been possible due to the introduction of large-scale datasets required by the data hungry DL algorithms, especially for domain specific problems such as road damages, as the dataset introduced by the University of Tokyo [22]. This dataset contains types of structural damages (cracks, patches, etc.), but it is still limited, as it presents unbalanced classes (specifically for potholes, which are quickly repaired in Japan and thus unrepresented). Thus, one of the contributions of this article is to complement the original dataset with more samples per class, in particular for the potholes and alligator cracks classes, which are very common in tropical regions and countries like Mexico and thus essential for a great variety of case uses.

In the following sections we will describe the dataset, the initial experiments performed with various algorithms and we will discuss how they compared with the state of the art described above, outlining as well future avenues of research.

# 4 Proposed Solution

Considering the progresses made in the domain, the integration of the technologies described above into "smart cities" paradigms such as digital asset management systems was only question of time, and some government and private service companies have started to make use of the information collected into road damage databases in various ways. The main idea is to leverage the progresses made in various domains (AI, IoT, mobile and cloud computing) for creating systems that can ease the labor of road inspectors on the one hand, but which can also be used to implement end-to-end solutions for creating a large database of structural damages from individual roads, to streets in medium to large cities, leading to the so-called "digital twins" (**Figure 4** b)).

These systems could aid in improving the work of the municipalities or governments at the federal level, helping them in managing their assets. Examples of platforms with these capabilities have been started to be deployed in the US (RoadBotics [23]), but their service is more oriented towards a more general assessment of the street qualities.

The main idea of our work is to leverage the power of AI algorithms in tandem with mobile and cloud computing to propose a system that can acquire geo-referenced images of road damages, in order to create a "digital map", which can be used for many other purposes, such as big data analyses. In this paper we focus on the creation of the dataset and training of the acquisition system, other components of the system will be described in subsequent work.

## 4.1 Utilized dataset

As mentioned previously, the research on road damage detection suffered from a lack of unified datasets. This problem has been partially addressed by the community, with various datasets of various sizes and characteristics; for instance, researchers at the Universita di Roma Tre introduced a dataset [16] with three major structural damage classes (linear and alligator cracks and potholes) and used it for implementing an acquisition system using a mobile phone. More recent efforts have sought to include a broader category of classes, such as the dataset of University of Tokyo [22], which includes 8 types of damages (images of 600x600 pixels), as depicted on **Table 1**.

**Table 1.** Road damage types in the dataset proposed in [21] and their definitions

| Damage Type | | | Detail | Class Name |
|---|---|---|---|---|
| Crack | Linear Crack | Longitudinal | Wheel mark part | D00 |
| | | | Construction joint part | D01 |
| | | Lateral | Equal interval | D10 |
| | | | Construction joint part | D11 |
| | Alligator Crack | | Partial pavement, overall pavement | D20 |
| Other Corruption | | | Rutting, bump, pothole, separation | D40 |
| | | | White line blur | D43 |
| | | | Cross walk blur | D44 |

The codes in **Table 1** are used by the Japanese Government to classify road damages and it is used by the authors to categorize structural damages in their work. This dataset is the largest currently available and has been extensively used to implement road inspection systems using deep learning architectures, such as SSD like RetinaNet [20] and MobileNet [21]. Although this dataset is relatively large, some classes or damage instances are poorly represented, such as the potholes class, which stems from the fact that these damages are quickly repaired in Japan.

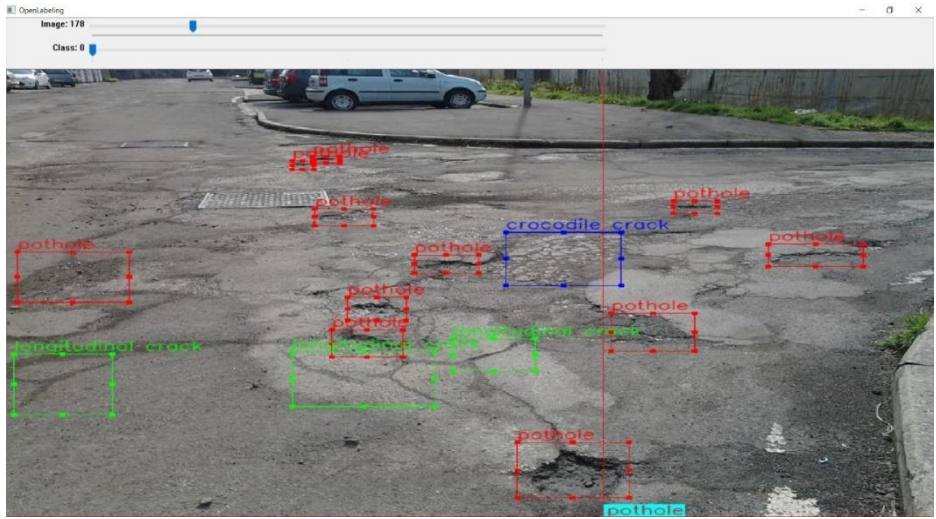

**Fig. 5.** An example of a training instance in our dataset, labels were manually annotated using Open Label Master, note the prominence of the pothole class in this image.

However, potholes at various stages of deterioration are one of the most important structural damages that need to me tracked by governmental bodies, and as such, one of the contributions of our work is to include more instances to the classes in the dataset introduced in [21] to have a more balanced representation, as it can be observed in **Table 2**. An instance of the enlarged dataset, containing multiples class labels in shown in **Figure 5**. We included more instances of the D00, D10 and D11 classes (longitudinal cracks of various types) and for the D40 (potholes), for which several hundred examples have been added, as shown on the third row of **Table 2**.

**Table 2.** Number of damage instances in each class for the original dataset presented in [21] and the one used in this work. Note that the potholes and cracks classes are better represented.

| Dataset/Class | D00 | D01 | D10 | D11 | D20 | D40 | D43 | D44 |
|---|---|---|---|---|---|---|---|---|
| Maeda [21] | 2,678 | 3,789 | 742 | 636 | 2,541 | 409 | 817 | 3,733 |
| Modified | **2,978** | 3,789 | **1042** | **1036** | 3,341 | **1,609** | 817 | 3733 |

Using this extended or modified dataset, we have carried out a comparative analysis with previous works in the literature, in order to evaluate if any gains in performance could be achieved, which in most instances was the case. In the next section, we will discuss how the dataset was used for training a set of "generic object detectors", which are the backbone for our acquisition and MMS system. It must be noted that most of the images are taken from "above the road" using mobile devices.

The choice of acquisition system was justified in **Section 2** from the "technical" perspective, but it should be stressed that in many countries, installing a camera outside the car constitutes a violation of the law [22]. Thus, a great number of solutions are based on inexpensive mobile phone cameras have been proposed and, as mentioned in **Section 3**, they require small footprints and real-time performances, whilst attaining high detection and classification accuracies.

We have discussed already the advantages of using various recently developed "generic object detection" based on deep learning architectures; in what follows, we will discuss how we trained various of these algorithms, later we will compare them in terms of performance.

## 4.2   Training

For the sake of completeness, we have implemented three types of object detectors, concentrating on those amenable for implementation on mobile and embedded platforms: an LBP-cascade classifier (as in [16]) as well more recent, DL-based generic object detectors, specifically RetinaNet (as in [20]) and MobileNet (as in [22])

In the case of RetinaNet, and as discussed in [20], this model can be trained with different backbone neural networks for the feature extraction phase (RestNet or VGG in its various configurations), trading accuracy versus inference time, obtaining in general better results than other popular Single Shot Detectors (SSD) such as YOLO, whilst yielding models amenable for embedded or mobile implementations. For instance, RetinaNet with VGG19 as backbone requires 115.7MB and can achieve inferences of 0.5s, fast enough for most applications in this domain.

We carried out the training of the deep learning models described above using the following methodology. The labeled dataset is randomly shuffled and separated into training set and validation set, 30% percent of the images of each class was used for training, taking care that the model did not overfit; for this purpose we have made use of dropout and data augmentation over the original dataset (using various image transformations such as cropping, warping, rotation, among other)

Data augmentation is a method for "synthetically" generating more training data from the existing training examples, by submitting the latter to a variety of random transformations that yield believable looking images. The process of data augmentation can reduce significantly the validation loss, as depicted in **Figure 6**, where the plots for the training and validation for two identical ConvNets are shown, one without using data augmentation and the other using the strategy.

Using data augmentation clearly combats the overfitting problem which can be observed in the left side of the figure (this is the validation loss decreases and then increases again, whilst the training loss decreases for the training set). The right side of the figure shows how the validation loss can be regularized using data augmentation, among other strategies.

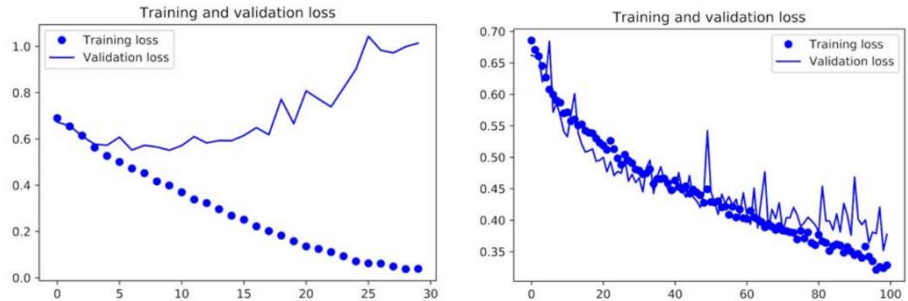

**Fig. 6.** Train and validation loss for 2 identical DNNs, a) without using data augmentation and b) data augmentation. The loss clearly shows how D.A prevents overfitting

The DNNs description as well as the training phase were implemented with Keras and Tensorflow running on Google with a Tesla K80 GPUs; the models were trained with batch size of 24 and number steps per epoch of 300, using Adam as optimizer, adapting the learning rate as discussed in [22]. As we will see in the next section, the proposed approach obtained improvements over previous works, especially for the underrepresented classes in the original road damages dataset.

## 5      Results and discussion

In our experiments, both the training and testing phases were carried out using Google Colab, which enables us to have access to high performance computing (a Tesla K80 GPU and other resources) and to host our DB on the Google Cloud, all while avoiding any computational burden. In the evaluation phase, we also made use of this tool, importing the trained models and the 30% of the validation set; furthermore, we carried out experiments on real-time video to test the capabilities of the proposed models in a variety of scenarios.

**Table 3** shows the obtained results for the three "generic object detectors" tested in our experiments. As mentioned before, we decided to implement an LBP-cascaded based classifier to contrast our results with those reported in [16], as we made use of a part of their dataset. We also report the results for the deep learning approaches reported in the literature, which make use of the RetinaNet [20] and MobileNet [22]. The proposed models achieved consistently better results than those reported in the literature using the same methods ([16], [20] and [22]) respectively, while achieving higher accuracies and inference time, as we will discuss later in this section.

**Table 3.** Number of damage instances in each class for the original dataset presented in [21] and the one used in this work. Note that the potholes and cracks classes are better represented.

| Metric/Class | D01 (Long Crack) | D11 (Lat. Crack) | D20 (Alligator) | D40 (Pothole) |
|---|---|---|---|---|
| LPB-cascade | 0.7080 | 0.7182 | 0.7625 | 0.7476 |
| MobileNet | 0.8124 | 0.9032 | 0.9234 | 0.9675 |
| RetinaNet | **0.9148** | **0.9511** | **0.9522** | **0.9823** |

In **Table 3**, we have decided to show the results only for 4 of the most reported classes (longitudinal and lateral cracks, alligator cracks and potholes), as they are the most prevalent and it makes the comparisons with previous work easier. It should be noted that other works discussed in the state of the art in **Section 3** have obtained good results for individual road damage classes (even with pixel-level accuracies) but they were not amenable to mobile or embedded vision implementations.

The sole work in the literature that so far has made a comprehensive analysis is the work of the researchers at the University of Tokyo [21], which implemented a couple of generic object detectors using Inception V2 and MobileNet. The authors obtained the best results with the latter, but more recent detectors have been proposed in the literature, such as RetinaNet, for which we obtained the best results, as depicted in **Table 3**.

In order to assess the improvements obtained by our extended dataset, we show a comparative analysis between the aforementioned MobileNet-based classification and detection method [21] and the best of our models, which is shown on **Table 4**. The SSD detector and classifier performed well, with both high precision and recall, especially compared with the D11 and D40 classes (lateral cracks and potholes, respectively). This can be attributed to the larger number of training instances in our dataset, a problem reported by the authors in [21]. The authors mentioned the class D43 (white line blurs) as a counter example, as the number of examples is limited, but we believe that this class is relatively easier to detect and classify, which might explain the results.

**Table 4.** Number of damage instances in each class for the original dataset presented in [22] and the one used in this work. Note that the potholes and cracks classes are better represented.

| Metric | Road Damage Class | | | | | | | |
|---|---|---|---|---|---|---|---|---|
| ***Maeda et al [21]*** | D00 | D01 | D10 | D11 | D20 | D40 | D43 | D44 |
| Recall | 0.40 | 0.89 | 0.20 | 0.05 | 0.68 | 0.02 | 0.71 | 0.85 |
| Precision | 0.73 | 0.64 | 0.99 | 0.95 | 0.68 | 0.99 | 0.85 | 0.66 |
| Accuracy | 0.81 | 0.77 | 0.92 | 0.94 | 0.83 | 0.95 | 0.95 | 0.81 |
| ***Best Model (ours)*** | | | | | | | | |
| Recall | 0.60 | 0.90 | 0.40 | 0.40 | 0.76 | 0.70 | 0.80 | 0.70 |
| Precision | 0.87 | 0.70 | 0.89 | 0.92 | 0.92 | 0.88 | 0.87 | 0.82 |
| Accuracy | 0.91 | 0.81 | 0.92 | 0.95 | 0.95 | 0.98 | 0.95 | 0.84 |

The accuracies obtained by our RetinaNet-based model in which, as in the case of the authors in [20], we have used VGG19 as the backbone network. However, the authors in this paper the authors make use of a single metric (a mAP of 0.8279) for reporting their result, without performing a per-class comparison.

In **Table 5**, we report some of the features of the resultant RetinaNet-based model, compared with other two state of the art models. First, the model size is relatively compact (125.5MB), while achieving a low inference time (0.5 s) and a higher mean average precision (mAP), making our proposal one of the more

**Table 5.** Metrics comparison for our best model against the state of the art

| Metric/Model | Model Size | Inference Time | Best mAP | Type of Device |
|---|---|---|---|---|
| RetinaNet [20] | 115.7 MB | 0.5 s | 0.8279 | Mobile |
| MobileNet [22] | N/A | 1.5 s | N/A | Mobile |
| RetinaNet (ours) | 125.5 MB | 0.5 s | 0.91522 | Mobile |

# 6 Conclusions and future work

In this study, we have presented the first block for a digital asset management tool, a generic object detector trained to identify and classify road damages from still images or real-time video with high accuracy and low inference time. We have tested different models and approaches to solve this problem, based on traditional computer approaches and on more recent deep learning architectures.

For the training of our models we deployed the relatively large dataset originally introduced by the University of Tokyo, augmenting it with images from Mexican and Italian roads, in order to compensate the disbalance present for some road damages such as potholes. The obtained results clearly show that RetinaNet can outperform other state of the art road damage detectors and it is amenable for a mobile or embedded implementation (i.e. for ADAS applications) while running at acceptable rates (0.5 seconds in our current implementation), which has been demonstrated to be more than enough for a vehicle with a speed of 40 km/s to avoid information leakage or duplication, as reported in [22] for an inference time of 1500 ms.

As future work, we plan to integrate the model running on a mobile device to a full-fledged Digital Asset Management Asset platform running on Amazon Web Services. The main idea is to leverage the capabilities of the cloud computing system for creating and maintaining a digital twin of the roads of a region or the streets of the city described in **Section 2**. This geo-localized database could be used for monitoring and prognosis purposes, if coupled with a predictive model, leading to a more informed process in the allocation of resources for maintenance, as well as the required periodicity

# Acknowledgments

We thank Prof. Benedetto from University de Roma Tre (Italy) for kindly providing some of the images used to complement the dataset from the University of Tokyo (Japan), along other images obtained by our group in Guadalajara, Mexico. We also thank the people of Vidrona LTD (Edinburgh, UK) for helping us to delimit the problem addressed in this paper, especially to Shailendra and Ashutosh Natraj.

1